\documentclass{article}

     \PassOptionsToPackage{numbers, compress}{natbib}

     \usepackage[preprint]{neurips_2020}

\usepackage[utf8]{inputenc} 
\usepackage[T1]{fontenc}    
\usepackage{url}            
\usepackage{booktabs}       
\usepackage{amsfonts}       
\usepackage{nicefrac}       
\usepackage{microtype}      

\usepackage{amsmath}
\usepackage{amssymb}
\usepackage{mathtools}
\usepackage{amsthm}
\usepackage{mathrsfs}
\usepackage{enumitem}
\usepackage{subfig}

\newtheorem{theorem}{Theorem}

\title{Reinforcement Learning with Feedback-modulated TD-STDP}

\author{%
  Stephen Chung \\
  Department of Computer Science\\
  University of Massachusetts Amherst\\
  Amherst, MA 01003 \\
  \texttt{minghaychung@umass.edu} \\
  \And
  Robert Kozma\\
  Department of Computer Science\\
  University of Massachusetts Amherst\\
  Amherst, MA 01003 \\
  \texttt{rkozma55@gmail.com} \\
}

\begin{document}

\maketitle

\begin{abstract}
Spiking neuron networks have been used successfully to solve simple reinforcement learning tasks with continuous action set applying learning rules based on spike-timing-dependent plasticity (STDP). However, most of these models cannot be applied to reinforcement learning tasks with discrete action set since they assume that the selected action is a deterministic function of firing rate of neurons, which is continuous. In this paper, we propose a new STDP-based learning rule for spiking neuron networks which contains feedback modulation. 
We show that the STDP-based learning rule can be used to solve reinforcement learning tasks with discrete action set at a speed similar to standard reinforcement learning algorithms when applied to the CartPole and LunarLander tasks. Moreover, we demonstrate that the agent is unable to solve these tasks if feedback modulation is omitted from the learning rule. We conclude that feedback modulation allows better credit assignment when only the units contributing to the executed action and TD error participate in learning. 
\end{abstract}


\section{Introduction}
In recent years, spiking neural networks (SNNs) gained popularity in solving machine learning tasks \cite{pfeiffer2018deep, huh2018gradient, zhang2019spike, obeid2019structured}. Still, recent advances in deep learning mostly focus on artificial neural networks (ANNs) instead of SNNs, partly because ANN can be trained efficiently by back-propagation, while SNNs have problems with back-propagation since their units communicate by binary spikes, therefore the network is non-differentiable \cite{dayan2001theoretical}. However, back-propagation is generally believed to be biologically implausible \cite{hassabis2017neuroscience}. 
However, spike-timing dependent plasticity (STDP) and reward-modulated STDP (R-STDP) have been experimentally observed in biological neural systems \cite{gerstner2014neuronal}. STDP can be derived theoretically as a result of maximizing mutual information between presynaptic neuron and postsynaptic neuron \cite{toyoizumi2005spike}, while R-STDP can be derived theoretically as a result of maximizing a global reward signal \cite{florian2007reinforcement, fremaux2013reinforcement}. 
  
Previous works have shown some success in solving reinforcement learning tasks with SNN using STDP-based learning rules. For tasks with continuous action set,  \cite{vasilaki2009spike, fremaux2013reinforcement} showed that R-STDP or TD-error-modulated STDP (TD-STDP), a learning rule similar to R-STDP but with reward replaced with temporal difference error (TD error), can be used to solve reinforcement learning tasks with continuous action set such as maze task and inverted CartPole. \cite{bing2018end} showed success in solving a lane-keeping task using R-STDP. 
 
For tasks with discrete action set, \cite{aenugu2019reinforcement} used a population of SNNs to successfully learn tasks such as Mountain Car and CartPole, but the learning is much slower than that of standard reinforcement learning algorithms. \cite{potjans2009spiking} solved the task of grid-world successfully, but their algorithm could only apply in tasks with finite and moderate number of states. \cite{wunderlich2019demonstrating} used R-STDP to solve a simplified version of Pong. 
  
In these SNN models (except \cite{aenugu2019reinforcement}), the action chosen is a deterministic function of firing rate of actor neurons, therefore they can only rely on the stochastic activity of spike train to encourage exploration. However, this way of encouraging exploration is inefficient since it is difficult for all neurons to have their independent random noise aligned. For instance, when using the average of a group of neurons' firing rate as strength of action, the random noise necessary for exploration will mostly cancel out after taking the average. This method of exploration works only if the action set is continuous and the reward function is continuous function of action, since a very small random noise in action is sufficient to estimate change in return, which is essentially equivalent to gradient estimation by numerical approximation. If the action set is discrete and a threshold function is used to convert firing rate to action, the gradient will be zero almost everywhere and there will be no exploration if noise is too small. 
  
  Actor-critic algorithms are one of the earliest and most popular methods investigated in reinforcement learning \cite{witten1977adaptive, barto1983neuronlike}. The actor learning rule of actor-critic algorithm is similar to long-term potentiation (LTP) side of R-STDP, but with state replaced by traces of incoming spikes in SNN. It has also been hypothesized that actor and critic network may correspond to dorsal and ventral subdivision of the striatum in the biological neural system \cite{takahashi2008silencing}, and TD error representing dopamine modulation in biological neural systems \cite{glimcher2011understanding}.
  
Recent neuroscience studies indicate that feedback connections may have roles in learning and attention. \cite{roelfsema2018control} summarized different evidences indicating the presence of feedback modulation in learning synaptic weight, and stated a gating hypothesis to explain the evidences, which suggests that ``\textit{response selection elicits feedback signals that enable the plasticity of upstream synapses}''. The feedback-modulated TD-STDP learning rule we propose corresponds to this hypothesis. \cite{roelfsema2005attention, rombouts2015attention} also proposed learning rule with feedback modulation for rate-based neurons. As the main goal of this paper is to solve reinforcement learning task with SNN, we refer readers to \cite{roelfsema2018control} for relevant works on feedback modulation in neuroscience.
  
In this paper, 
we investigate how to solve a reinforcement learning task efficiently with discrete action set using SNN and STDP-based learning rules. We propose an actor-critic architecture that treats actor neurons' output as the probability of choosing an action instead of the action itself. Since the action chosen is no longer a deterministic function of actor neurons' output, we also propose a new learning rule, which we called feedback-modulated TD-STDP, to allow better credit assignment. 
We show that the new feedback-modulated TD-STDP learning rule can be used to solve common reinforcement learning tasks such as CartPole and LunarLander at a speed similar to standard reinforcement learning algorithms. At the same time, TD-STDP and R-STDP are unable to learn the task without feedback modulation.

It is important to point out that a recent work, developed independently from us, proposed a learning rule similar to ours, called e-prop \cite{bellec2020solution}. It is based on back-propagation through time (BPTT) and the paper showed that it can solve some Atari games efficiently. Their proposed learning rule for solving reinforcement learning tasks with LIF neurons have several differences from our approach: (i) we use LTD in STDP while they do not use such; (ii) they use different approach in the computation of pre-synaptic traces; (iii) we use regularization which is needed for more complex tasks. Our paper was submitted to a peer-reviewed conference before \cite{bellec2020solution} was published and we were unaware of this study before submission of our paper. Apart from \cite{bellec2020solution}, we are not aware of any prior works that uses SNN with STDP-based learning rule to solve common reinforcement learning tasks with discrete action set at a speed similar to standard reinforcement learning algorithms. 

The results introduced in this paper open the prospect of broader application of SNNs in combination with reinforcement learning to solve machine learning problems efficiently.

\section{Background}

\subsection{Markov Decision Process}

We will consider a continuous-time Markov Decision Process (MDP); for detailed definition, see \cite{guo2009continuous}. A continuous-time MDP is a tuple given by $(\mathcal{S}, \mathcal{A}, q(j|i,a), r(i,a))$, where:

$\mathcal{S}$ is the set of possible states of the environment. The state at time $t$ is denoted by $S_t$ which is in $\mathcal{S}$;

$\mathcal{A}$ is the set of possible actions. The action at time $t$ is denoted by $A_t$ which is in $\mathcal{A}$. We only consider discrete action set here;

$q(s'|s,a)$ is the transition rate function which describes dynamics of state;

$R(s,a)$ is the reward function, defined to be $E[R_t|S_t=s, A_t=a]$, where $R_t$ is the reward at time $t$ which is in $\mathbb{R}$.

We are interested in finding a policy function $\pi: (\mathcal{S}, \mathcal{A}) \rightarrow \mathbb{R}$ such that if action is sampled from policy function, that is, if $A_t|S_t \sim \pi(S_t, \cdot)$, then the expected return $E[G_t|\pi]$ is maximized, where return is defined to be $G_t := \int_{s=t}^{\infty} \exp\left(\frac{-(s-t)}{\tau}\right) R_s ds$. $\tau$ is the time constant for discount and $\exp\left(\frac{-t}{\tau}\right)$ is the analog of discount factor $\gamma^t$ in discrete case. Value function is defined to be $V^{\pi}(s) := E[G_t|S_t=s, \pi]$.

\subsection{Spiking Neural Network (SNN)}

For the SNNs employed here, all neurons use standard leaky-integrate-and-fire (LIF) model \cite{dayan2001theoretical} with no refractory period. The membrane voltage $V_j$ of neuron $j$ is computed as:
\begin{equation}
\tau \frac{dV_j(t)}{dt} = E_{res} - V_j(t) + R \sum_i w_{ij} X_i(t) \label{eq:s0}
\end{equation}

where $\tau$ is the time constant of the neuron, $E_{res}$ is the resting potential, $R$ is the resistance, $X_i(t)$ is the spike train from presynaptic neuron $i$ and $w_{ij}$ is the synaptic weight from neuron $i$ to neuron $j$. Neuron $j$ fires if $V_j(t) > \theta$ where $\theta$ is the threshold for firing. Immediately after firing, $V_j(t)$ is reset to $E_{res}$. We use $E_{res}=-65$mV, $\theta=-52$mV, $\tau=100$ms and $R=1 \Omega$ in our model.

\section{Model}
\begin{figure}
  \centering
  \includegraphics[width=0.7\textwidth]{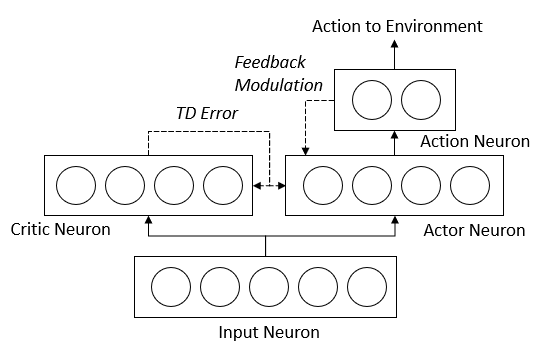}
  \caption{Architecture of the proposed actor-critic model. The environment state is converted to spikes at the input neurons. The spikes are passed from input neurons to both critic and actor neurons. Firing rate of actor neurons will determine the firing probability of action neurons, which encode the action directed to the environment.}
   \label{fig:7}  
\end{figure}

The proposed model architecture is shown on Fig \ref{fig:7}, which is based on actor-critic architecture. We will discuss each group of neurons in the following sections.

\subsection{Critic Neuron}
The role of critic neurons is to estimate the value function of the state, $V(S_t)$. The architecture of the critic neurons largely follows that of \cite{fremaux2013reinforcement} except a slight change in computation of TD error in discrete step, which is explained in the Appendix.

To learn value function, we can minimize TD error of the current value estimation. Denote $\hat{V}(s)$ as the estimated value function, then continuous version of TD error is given by (same as \cite{fremaux2013reinforcement}; here $\hat{V}'(S_t)$ denotes the derivative of $\hat{V}(S_t)$ w.r.t. time $t$):
\begin{equation}
\delta(t) = \hat{V}'(S_{t}) - \frac{1}{\tau}\hat{V}(S_{t}) + R(t) \label{eq:0}
\end{equation}

Estimation of value is based on firing rate of critic neurons. We first compute firing rate of a single neuron using temporal average, given by:
\begin{align}
\rho_j(t) &= \frac{1}{\tau_n} \int_{-\infty}^t \exp\left(-\frac{t-s}{\tau_n}\right)X_j(s)ds  \label{eq:9} 
\end{align}
where $\rho_j(t)$ is the firing rate of neuron $j$, $X_j(s)$ is the spike train of neuron $j$ and $\tau_n$ is the time constant for the temporal filter. 

Then, assuming there are $N_e$ excitatory critic neurons and $N_i$ inhibitory critic neurons, the estimated $\hat{V}(S_{t})$ is given by a linear function of average firing rate of excitatory critic neurons subtracting the average firing rate of inhibitory critic neurons:
\begin{align}
\hat{V}(S_t) &=\alpha \left(\frac{1}{N_e} \sum_{j=1}^{N_e} \rho^c_{e,j}(t) - \frac{1}{N_i} \sum_{j=1}^{N_i} \rho^c_{i,j}(t)\right) + \beta \label{eq:5}
\end{align}

where $\rho^c_{e,j}(t)$ is the firing rate of excitatory critic neurons $j$, $\rho^c_{i,j}(t)$ is the firing rate of inhibitory critic neurons $j$, $\alpha$ and $\beta$ are scalar constant for linear transformation.

\subsection{Learning Rule for Critic Neuron}

When the TD error is positive, it is expected that the excitatory critic neurons fire more rapidly to reduce the TD error, and vice versa. Thus, we can treat TD error as reward for the excitatory critic neurons. Similarly, we can use negative of TD error as reward for the inhibitory critic neurons. Therefore, we can use R-STDP with eligibility trace\footnote{R-STDP with eligibility trace discussed here is equivalent to MSTDPET from \cite{florian2007reinforcement}}, which is shown theoretically that can maximize reward \cite{florian2007reinforcement}, to train critic neurons. 
In this case, however, the reward is replaced by TD error. The learning rule of weight from input neuron $i$ to critic neuron $j$ is then:
\begin{align}
\frac{d w_{ij}(t)}{dt} &= \pm \eta \delta(t) z_{ij}(t) \label{eq:8} \\
\frac{d z_{ij}(t)}{dt} &= -\frac{z_{ij}(t)}{\tau_z} + A_+ P_{i}(t) X_j(t) - A_- P_{j}(t) X_i(t)  \\
\frac{d P_{i}(t)}{dt} &= -\frac{P_{i}(t)}{\tau_p} + X_i(t) \\
\frac{d P_{j}(t)}{dt} &= -\frac{P_{j}(t)}{\tau_p} + X_j(t) \label{eq:8E}
\end{align}
where $\eta$ is the learning rate, $z_{ij}$ is the eligibility trace, $P_{i}(t)$ and $P_{j}(t)$ are the trace of presynaptic spike train and postsynaptic spike train respectively, $A_+$ and $A_-$ are constants determining the strength of LTP and LTD of STDP respectively, $\tau_p$ and $\tau_z$ are time constant, and $\delta(t)$ is the TD error computed using (\ref{eq:0}). 

In (\ref{eq:8}), excitatory critic neurons are updated with positive sign while inhibitory critic neurons are updated with negative sign. It can be shown that the above learning rule approximately follows a gradient descent on the squared TD error, which is the critic's learning rule in actor-critic model \cite{sutton2018reinforcement}. This is stated in the follow:

\begin{theorem}\label{thm:1}
Assume that $A_+ = 1$, $A_- = 0$, $\tau_p = \tau$ and $\tau_z = \tau_n$.
Then the learning rules given by (\ref{eq:8}) to (\ref{eq:8E}) are approximated as follows:
\begin{align}  \frac{dw_{ij}(t)}{dt}   &\approx \hat{\eta} \delta_t \nabla_{w_{ij}} \hat{V}(S_t),  
\end{align}
where $\hat{\eta}$ is a constant, and the approximation is given by the straight-through estimator (STE) \cite{bengio2013estimating}.
\end{theorem}

The detailed proof can be found in Appendix.

Comments:
\begin{enumerate}
\item The approximation is due to use of straight-through estimator (STE) \cite{bengio2013estimating}, since the step function in determining firing of neurons has zero derivative almost everywhere. 
\item
The update rule in (\ref{eq:8}) can be further combined with Adam optimizer \cite{kingma2014adam}, by treating the update as gradient input to Adam optimizer. We found that this stabilizes the learning process as the use of momentum smooths out temporal noise.
\end{enumerate}

\subsection{Actor and Action Neuron}

As a novel aspect of this approach, we treat the output of the actor neurons as the probability of choosing an action instead of the action itself. To be specific, let assume that the action set has $K$ discrete actions, and we have $N_e$ excitatory actor neurons and $N_i$ inhibitory actor neurons per discrete action, making a total of $N = K (N_e + N_i)$ actor neurons. We compute the firing rate of each actor neuron using the same formula in (\ref{eq:9}), and the average firing rate of excitatory actor neurons and inhibitory actor neurons for action $k$, denoted by $\rho^a_{e, k} (t)$ and $\rho^a_{i, k} (t)$ respectively, is given by:
\begin{align}
\rho^a_{e, k} (t) &= \frac{1}{N_e} \sum_{j=1}^{N_e} \rho^a_{e, k, j}(t) \quad \text{for } k\in 1, 2, ...K \label{eq:AN_1}\\
\rho^a_{i, k} (t) &= \frac{1}{N_i} \sum_{j=1}^{N_i} \rho^a_{i, k, j}(t) \quad \text{for } k\in 1, 2, ...K
\end{align}
where $\rho^a_{e, k, j}(t)$ and $\rho^a_{i, k, j}(t)$  is the firing rate of excitatory actor neuron $j$ and inhibitory actor neuron $j$ for action $k$ respectively. Then the final action at time $t$, $A(t)$, is chosen according to a softmax function of the average firing rate of actor neurons:
\begin{align}
s(t) :&= softmax (\alpha(\rho^a_e (t)- \rho^a_i (t)))\\
P(A(t) = k) &= s_k(t)  \label{eq:7}
\end{align}
where $\rho^a_e (t) = [\rho^a_{e,1} (t), \rho^a_{e,2} (t), ..., \rho^a_{e,K} (t)]^T$,  $\rho^a_i (t) = [\rho^a_{i,1} (t), \rho^a_{i,2} (t), ..., \rho^a_{i,K} (t)]^T$, $s(t)$ is the vector of action probability, and $\alpha$ is the temperature constant to control the degree of exploration. As $\alpha$ becomes larger, the model will degenerate to the deterministic case where there is no exploration. As $\alpha$ approaches $0$, the model will just degenerate to a uniform policy. This method thus allows effective coordination of exploration across a large number of action units. To prevent rapid oscillation of $A(t)$, we can re-sample this every $m$ time step instead of every time step.

We can also think of $A(t)$ having a corresponding one-hot representation in a layer on top of actor neurons and we call it action neurons, which corresponds to the final action chosen according to (\ref{eq:7}). Thus, there will be a total of $K$ actor neurons, one per action, whose firing corresponds to the action being executed at the moment. That is, the spike train of action neurons are $A_k(t) = I\{A(t)=k\}$. 

\subsection{Learning Rule for Actor Neuron}

In theory, the critic learning rule above can be directly applied to actor neuron and it still maximizes the return as shown by \cite{fremaux2013reinforcement}. However, this is very inefficient in credit assignment if the action chosen is a stochastic function of firing rate of actor neurons. Consider the case of only two actions, where the two groups of actor neurons are firing at the same rate, and one action is then randomly chosen. A positive TD error is received after the action. If we continue to use the critic learning rule, then both groups will be rewarded, even though only the chosen action is causing the positive TD error. A more efficient credit assignment method is to assign credit only to the actor neurons causing the chosen action, that is, the learning rule has to be regulated by $A(t)$, as in the three-factor learning rule in actor-critic model \cite{sutton2018reinforcement}.

In light of this idea, we propose a new learning rule that is modulated by $A(t)$, that is, there is a feedback connection from action neurons that modulate the learning. Therefore, we call this learning rule feedback-modulated TD-STDP. The proposed learning rule for weight from input neuron $i$ to actor neuron $j$ (denote $k$ as the action for this actor neuron) is as follows:
\begin{align}
\frac{d w_{ij}(t)}{dt} &= \pm \eta \delta(t) q_{ij}(t) \label{eq:6}\\
\frac{d q_{ij}(t)}{dt} &= -\frac{q_{ij}(t)}{\tau_q} + (A_k(t) - s_k(t))\ z_{ij}(t) \label{eq:3}\\
\frac{d z_{ij}(t)}{dt} &= -\frac{z_{ij}(t)}{\tau_z} + A_+ P_{i}(t) X_j(t) - A_- P_{j}(t) X_i(t)  \label{eq:10}\\
\frac{d P_{i}(t)}{dt} &= -\frac{P_{i}(t)}{\tau_p} + X_i(t) \\
\frac{d P_{j}(t)}{dt} &= -\frac{P_{j}(t)}{\tau_p} + X_j(t) \label{eq:6E}
\end{align}
where $A_k(t)$ is the spike train for action neuron, $q_{ij}$ is a feedback-gated trace and $\tau_q$ is the time constant for the trace. Other notations are the same as the notations in learning rules of critic neurons. In (\ref{eq:6}), excitatory actor neurons are updated with positive sign while inhibitory actor neurons are updated with negative sign. 

As compared it with the critic neuron learning rule, the only difference in the actor neurons is the addition of (\ref{eq:3}), which gates the eligibility trace by feedback signal $A_k(t) - s_k(t)$. With this feedback gate, if TD error is positive, only the actor neurons corresponding to the chosen action will be rewarded while all other actor neurons will be punished (the case for negative TD error is similar). One may also wonder why $s_k(t)$ is in the learning rule. This actually corresponds to derivative of log of chosen action's probability:
\begin{equation}
\nabla_{\rho^a_{e, k}(t)} \log P(A(t)) = \alpha(A_k(t) - s_k(t))
\end{equation}

If we omitted $s_k(t)$, the agent is still learning in our experiment but the result is worse. The weight will easily explode to a very large magnitude. 

It can be shown that the above learning rule approximately equals to the actor's learning rule in actor-critic model with eligibility trace \cite{sutton2018reinforcement}:

\begin{theorem}\label{thm:2}
Assume $A_+ = 1$, $A_- = 0$, $\tau_p = \tau$, $\tau_z = \tau_n$ and $\tau_q =  \frac{1}{1 - \gamma \lambda}$. Then the learning rules given by (\ref{eq:6}) to (\ref{eq:6E}) for excitatory actor neurons are approximated as follows:
\begin{align}  
\frac{d\hat{q}_{ij}(t)}{dt} &\approx -(1- \gamma \lambda ) \hat{q}_{ij}(t) +  \nabla_{w_{ij}} \log P(A(t)|S(t))  \label{eq:th2_1}\\
\frac{dw_{ij}(t)}{dt} &= \hat{\eta} \delta(t) \hat{q}_{ij}(t) \label{eq:th2_2}
\end{align}
where $\hat{\eta}$ is a constant, and the approximation is given by the straight-through estimator (STE) \cite{bengio2013estimating}.
\end{theorem}

Again, the approximation is due to use of straight-through estimator (STE) \cite{bengio2013estimating}. For inhibitory actor neurons, a negative sign has to be placed on right-hand-side of (\ref{eq:th2_1}) and (\ref{eq:th2_2}). The detailed proof can be found in Appendix.

\subsection{Regularization in SNN}\label{sec:reg}

\subsubsection{Entropy Regularization}
Similar to ANN, entropy regularization can be employed to encourage exploration by preventing saturation of actions' probability \cite{mnih2016asynchronous}. In our model, gradient of entropy w.r.t weight from input neuron $i$ to excitatory actor neuron $j$ (denote $k$ as the action for this actor neuron) can be computed as:
\begin{align}
&- \frac{\partial}{\partial w_{ij}} \sum_{m=1}^{K}  P(A(t)=m) \log P(A(t)=m) \\=& - \sum_{m=1}^K s_m(t) (\log s_m(t) + 1)(I\{m=k\} - s_k(t)) \frac{\partial}{\partial w_{ij}} \alpha (\rho^a_{e,k} (t)- \rho^a_{i,k} (t)))\\
=& -\alpha \sum_{m=1}^K s_m(t) (\log s_m(t) + 1)(I\{m=k\} - s_k(t)) \frac{\partial}{\partial w_{ij}} \rho^a_{e,k} (t) \\
\approx & \frac{R \alpha }{N_e \tau_n \tau} (g_k \cdot ((X_j \cdot (X_i * k_{\tau})) * k_{\tau_n})) (t) \label{eq:r1}
\end{align}
where $g_k(t) = -\sum_{m=1}^K s_m(t) (\log s_m(t) + 1)(I\{m=k\} - s_k(t))$ and (\ref{eq:r1}) uses similar step in Appendix A.1. Absorbing $\frac{R \alpha }{N_e \tau_n \tau}$ into constant $c_e$ and assuming $A_-=0$, the learning rule in (\ref{eq:6}) is modified to\footnote{For the case of $A_- > 0$, one can use separate eligibility trace for LTP and LTD in (\ref{eq:10}), and use the eligibility trace for LTP to compute $(X_j \cdot (X_i * k_{\tau}))(t)$.}:

\begin{equation}
\frac{d w_{ij}(t)}{dt} = \eta ( \pm  \delta(t) q_{ij}(t) \pm c_e g_k(t) z_{ij}(t))\label{eq:11}\\
\end{equation}

In (\ref{eq:11}), excitatory neurons are updated with positive sign while inhibitory neurons are updated with negative sign.

In our experiment, we found that entropy regularization is necessary to solve LunarLander. Without entropy regularization, action 0 (doing nothing) will have almost zero probability after the first few hundreds of episodes, making the agent stuck in local optima policy.

\subsubsection{Weight Decay}
Weight decay, or L2 weight regularization, is also beneficial in training SNN. Since we do not employ any restrictions on weight's norm in our model, we observe that some actor neurons are firing on almost every step. But the absolute level of firing rate of actor neurons is unimportant. Only the relative level of firing rate of actor neurons determines the probability of choosing an action. By firing on every step, the above learning rule can no longer further increase the firing rate of a neuron. Therefore, a form of regularization is required to keep the absolute level of firing rate of actor neurons low such that the above learning rules are effective in controlling firing rate. We found that the use of weight decay can achieve such regularization. Denote $c_w$ as strength of weight decay, the learning rule in (\ref{eq:6}) is modified to:

\begin{equation}
\frac{d w_{ij}(t)}{dt} = \eta ( \pm  \delta(t) q_{ij}(t) - \frac{1}{2}c_w w_{ij}(t))\\
\end{equation}

\subsubsection{Target Firing Rate} \label{sec:tar}
Another mechanism to control firing rate of actor neurons is to do gradient descent on the squared difference between average firing rate of actor neurons $\rho^a_e(t) = \frac{1}{K}\rho^a_{e,k}(t)$ and a target firing rate $\hat{\rho}^a_e$. In our model, gradient of $(\rho^a_e(t) - \hat{\rho}^a_e)^2$ w.r.t weight from input neuron $i$ to excitatory actor neuron $j$ (denote $k$ as the action for this actor neuron) can be computed as:

\begin{align}
&-\frac{\partial}{\partial w_{ij}}(\rho^a_e(t) - \hat{\rho}^a_e)^2  \\
=& -\frac{2}{K}(\rho^a_e(t) - \hat{\rho}^a_e) \frac{\partial}{\partial w_{ij}} \rho^a_{e,k}\\
\approx & -\frac{2}{K} (\rho^a_e(t) - \hat{\rho}^a_e) \cdot (X_j \cdot (X_i * k_{\tau})) * k_{\tau_n}) (t) \label{eq:r2}
\end{align}

Again, (\ref{eq:r2}) uses similar step in Appendix A.1. Absorbing $\frac{2}{K} $ into constant $c_t$ and assuming $A_-=0$, the learning rule in (\ref{eq:6}) is modified to:

\begin{equation}
\frac{d w_{ij}(t)}{dt} = \eta ( \pm  \delta(t) q_{ij}(t) - c_t (\rho^a_e(t) - \hat{\rho}^a_e)z_{ij}(t))  \\
\end{equation}

The three methods of regularization proposed above can be combined as:

\begin{equation}
\frac{d w_{ij}(t)}{dt} = \eta ( \pm  \delta(t) q_{ij}(t)  \pm c_e g_k(t) z_{ij}(t) - \frac{1}{2}c_w w_{ij}(t) - c_t (\rho^a_e(t) - \hat{\rho}^a_e)z_{ij}(t))  \\
\end{equation}

\section{Experimental Results}

To test our proposed learning rule, we apply it to the CartPole problem (also called pole balancing and inverted pendulum) and LunarLander. Next, we will discuss the experimental details of the implementations. We used BindsNET \cite{Hazan2018} to simulate SNN in our experiment, with each step in SNN representing $1$ms.

For CartPole, we first perform Fourier transformation on input with a Fourier order of 2 \cite{konidaris2011value}, thus the output dimension is $81 = (2+1)^4$, with each output $o_i$ in the range of $[-1,1]$. Then we rescale the output to $[0,1]$ by $f_i= \frac{o_i+1}{2}$, which is the feature we used in CartPole. 

For LunarLander, we first perform Fourier transformation on input with a Fourier order of 1 and do not include any cross terms, thus the output dimension is $8$, with each output $o_i$ in the range of $[-1,1]$. Then we concatenate negative of these 8 outputs and a bias (a constant one) in the output vector, and finally apply ReLu on it to obtain the feature vector. The dimension of feature is thus 17, with each $f_i$ in the range of $[0,1]$. 

The firing rate per ms of input neurons is then given by the value of corresponding feature.  Since we use a time step of $1$ms in our stimulation of SNN, the firing rate is converted to spike by $X_i(t) \sim Ber(f_i(t))$ as input to the network. There are 1 input neuron per feature in CartPole and 16 input neurons per feature in LunarLander. We use the same features for baseline models, without conversion to spike.

We added a warm-up period of $100$ms before the start of each episode, where the initial state is presented to the agent without change for this period. No action has effects on state and no reward is given in this period. There is also no learning during the warm-up period. The reason for adding this period is to allow the firing rate of the agent to pick up from zero initially. 

\begin{table}
  \caption{Hyperparameters used in CartPole and LunarLander.}
  \label{table:1}
  \centering
  \begin{tabular}{lcccc}
    \toprule
         & \multicolumn{2}{c}{CartPole} & \multicolumn{2}{c}{LunarLander}\\
         \cmidrule(r){2-3}      \cmidrule(r){4-5}
         & Critic Network & Actor Network & Critic Network & Actor Network \\
    \midrule
		$N_e$ & 40 & 20 & 128 & 32 \\	    
		$N_i$ & 0 & 0 & 128 & 32\\	 
		$\eta$ & 2.5e-3 & 1e-2 & 1.25e-4 & 6.25e-5\\ 
		$\alpha$  & 2 & 25 & 4 & 15\\ 
		$\beta$  & -0.2 & n.a. & -2 & n.a.\\ 
		$\tau_n, \tau_p, \tau_z$ & 20 & 20 & 20 & 20\\ 
		$\tau_q$ & n.a. & 40 &  n.a. & 20\\	
		$c_e$ & n.a. & 0 &  n.a. & 1e-4\\			
		$c_w$ & n.a. & 0 &  n.a. & 1e-8\\	
		$c_t$ & n.a. & 0 &  n.a. & 0\\			
		$A_+$ & 1 & 1 & 1 & 1 \\				
		$A_-$ & 0 & 0 & 0 & 0 \\		
    \bottomrule
  \end{tabular}
\end{table}

The hyperparameters used in CartPole and Lunarlander experiment are shown on Table \ref{table:1}. We selected these hyperparameters values based on our educated guess followed by manual tuning to optimize performance. 

For CartPole, we scaled all rewards by 0.02, so the reward for each $1$ms before the end of episode is $0.001$. Time constant for discount rate is $1000$ms. We have not used batch update, inhibitory neurons, Adam optimizer, and any regularization. We re-sampled new actions at every environment step.

For LunarLandar, we scaled all rewards by $0.012$ and time constant for discount rate is $2000$ms. We used Adam optimizer with $\beta_1=0.995$ and $\beta_2=0.99995$, and a batch size of $16$. We re-sampled new action at every two environment steps.

Since one step in the environment represents multiple steps in SNN, we distribute the reward from the environment evenly to the corresponding SNN's time steps. We also set the target $\hat{V}(S_{t+\Delta t})$ to 0 in the last $2$ms of an episode when computing TD error.

\subsection{CartPole}

\begin{figure}%
    \centering
    \subfloat[CartPole]{{\includegraphics[width=.5\textwidth]{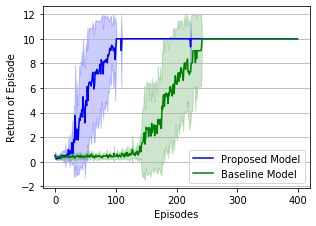} }}%
    \subfloat[LunarLander]{{\includegraphics[width=.5\textwidth]{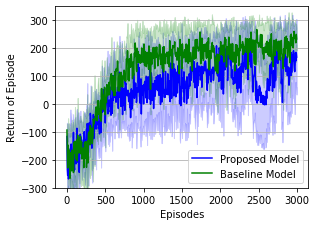} }}%
    \caption{Learning curve in proposed model and baseline model. Results are averaged over 10 runs, and shaded area represents standard deviation over the runs.}%
    \label{fig:10}%
\end{figure}

For CartPole, we used CartPole-v1 in OpenAI's Gym \cite{brockman2016openai} for our implementation of CartPole. Episode ends if it lasts more than 500 steps (equivalent to $10$s). The experimental results are shown in Fig \ref{fig:10}, which displays the episode return achieved by the agent for 400 episodes, averaged over 10 independent runs. Let ${T}_f$ denote the first episode when the agent achieves a perfect score (maintaining the pole for $10$s). The average value over 10 runs is $\bar{T}_f$ = 57.00, given in Table \ref{table:2}. In addition, let $\bar{T}_s$ denote the average number of episodes required for solving the task (defined as maintaining the pole for $10$s over all of the last 100 episodes); $\bar{T}_s$ = 169.50. 

\begin{table}
  \caption{Episodes required for learning the task.}
  \label{table:2}
  \centering
  \begin{tabular}{lcccccccccccc}
    \toprule
    & \multicolumn{4}{c}{CartPole} & \multicolumn{4}{c}{LunarLander}\\
     \cmidrule(r){2-5}      \cmidrule(r){6-9}
    & \multicolumn{2}{c}{${T}_f$} & \multicolumn{2}{c}{ ${T}_s$} & \multicolumn{2}{c}{${T}_f$} & \multicolumn{2}{c}{ ${T}_s$} \\
    \cmidrule(r){2-3}     \cmidrule(r){4-5}  \cmidrule(r){6-7}  \cmidrule(r){8-9}
    & Mean & Std. & Mean & Std. & Mean & Std. & Mean & Std. \\
    \midrule
Proposed  & 57.00 & 23.97 & 169.50 & 23.47 & 383.80 & 71.49 & 2575.20 & 666.51 \\ 
Baseline  & 180.70  &  23.85 & 301.90 & 28.97 & 503.30 & 76.03 & 1295.20 & 165.68\\ 
Model in \cite{aenugu2019reinforcement} &  1023.80 & 77.65 & \multicolumn{2}{c}{n.a.} & \multicolumn{4}{c}{n.a.} \\ 
    \bottomrule
  \end{tabular}
\end{table}

\begin{figure}[h!!!]
    \centering
	\includegraphics[width=.65\textwidth]{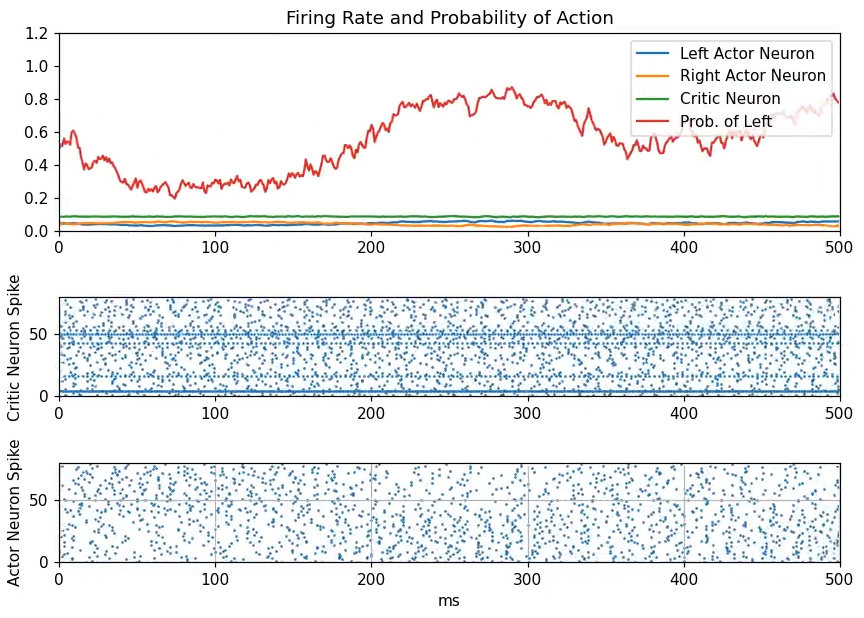}
	\caption{Illustration of the model after learning. First graph: Firing rate of both actor and critic neurons for a sample interval in an episode. Second and third graph: spike trains of critic and actor neurons. Actor neuron 1 to 40 is for left action and actor neuron 41 to 80 is for right action. The animation of this graph can be viewed online.}
    \label{fig:11}
\end{figure}

Next, we consider a comparison with other models. For baseline, we used a standard actor-critic model (one-step actor-critic (episodic) from \cite{sutton2018reinforcement}) with a similar architecture: both critic network and actor network are a one-hidden layer neural network with hidden layer's size of 40 using ReLu activation. This can be seen as the counterpart of our SNN model in ANN. Although this is not a state-of-the-art model, our goal is to show that SNN can be used to solve some common reinforcement learning tasks at a speed similar (or better) to their ANN counterpart. The results are shown in Table \ref{table:2}. It is seen that the proposed model is significantly better than the baseline considered. The learning speed obtained by feedback-modulated TD-STDP is on par with most standard reinforcement learning algorithms.

In addition, we have tested the model proposed in \cite{aenugu2019reinforcement}\footnote{In their paper, they used CartPole-v0 instead of CartPole-v1, which only differs in maximal length of episode. Maximal episode length of CartPole-v1 is 500 steps (10s) while that of CartPole-v0 is 200 steps (4s). We run their code published online on CartPole-v1 instead of CartPole-v0 in our experiment, using a population size of 10.}. Since their model does not converge for 400 episodes, we run the model for 2000 episodes instead. We noted that the model cannot solve the task (according to the definition above) in all 10 runs and $\bar{T}_f$, the average first episode agent achieves a perfect score, is 1023.80, significantly larger than both the baseline and our proposed model.

To illustrate the strength of the trained network, we have tested its performance over longer time intervals, i.e., not terminating the task after $10$s, which was used during training. Although the agent is trained for at most $10$s in any single episode, it is able to continue balancing the pole stably for a very long time beyond $10$s. In fact, we have to terminate its operation manually after it shows no signs of failing for more than 5 minutes. The video demonstrating this performance can be found on \url{https://youtu.be/Ti5CznX4H9I}.

As most hyperparameters are tuned to optimize performance instead of fitting biologically observed value, we find that the most neurons in above model are firing at a high frequency of above 200Hz, which is biologically implausible. However, this issue could be solved by tuning the relevant hyperparameters and using the regularization method discussed in Section {\ref{sec:reg}. For critic neurons, $\alpha$ could be set to a high value such that a low firing rate indicates a high value. For actor neurons, we can employ target firing rate discussed in Section \ref{sec:tar}. To be specifc, we used the same hyperparameters except for critic network: $\alpha=20$, $\beta=-1$, $N_e = 80$, and for actor network: $N_e=40$, $c_t=5\mathrm{e}{-6}$, $\bar{\rho^a} = 5\mathrm{e}{-3}$. After these changes, average firing rate of actor neurons and critic neurons are around 50Hz and 70Hz respectively, which are in the gamma frequency band, and $\bar{T}_f$ and $\bar{T}_s$ are $132.90$ and $251.00$ respectively, worse than the original model but still better than the baseline. At this firing rate, the dynamic of spiking neuron and rate-based neuron are also very different. The spike trains obtained by model simulations are illustrated in Fig \ref{fig:11}, and the animation of this could be found on \url{https://youtu.be/X7I6FpF3MJQ}.

We have tried to train the actor neuron with TD-STDP only without any feedback modulation. The agent fails to learn and episode length stays below 1s, showing that feedback modulation is necessary for the agent to learn.

\subsection{LunarLander}

We also applied the proposed model to LunarLander-v2 in OpenAI's Gym \cite{brockman2016openai}, a task which is much more complex than CartPole. There are four available actions in the task and the goal is to land the lander to the landing pad, which will yield 200 scores or above. The implementation of the model is largely the same as the one we used in CartPole, except: (i) we used 128 excitatory neurons and 128 inhibitory neurons in both critic network and actor network, (ii) we used Adam optimizer in updating weight, (iii) we added entropy regularization in learning, and (iv) we used a batch size of 16. 

We define perfect score as having a return of 200 or above, and solving the task as having an average return of 200 or above for the last 100 episodes. For each of the 10 independent runs, we run the model for 3000 or $T_s$ episodes, whichever is greater. The baseline model is similar to the one we used in CartPole, but with 128 hidden units for both critic and actor network.

The experimental result is shown in Fig \ref{fig:10}, which displays the episode return achieved by the agent for 3000 episodes, averaged over 10 independent runs. $\bar{T}_f$, the average value of the first episode that agent achieves a perfect score, is 383.80, and $\bar{T}_s$, the average value of the episodes required for solving the task, is 2575.20. A comparison with the baseline model is shown on Table \ref{table:2}. Though the proposed model is able to solve the task, the learning is slower than that of the baseline model. The performance of the feedback-modulated TD-STDP learning can be improved by optimizing its hyperparameters. Such optimization, however, has not been conducted extensively in this study. The present work aims at demonstrating the feasibility of the proposed approach for a more complex control problem.

We have made a video showing the trained model solving LunarLander, which could be found on \url{https://youtu.be/UzysCKolMFg}. All 10 episodes shown in the video achieve a score of over 200 (as long as the center of the lander is within the two flags, it is counted as successful landing). 

\section{Conclusions}

The goal of this paper is to to use SNNs to solve reinforcement learning tasks with discrete action set efficiently. In our approach, the actor-critic architecture treats actor neurons' output as the probability of choosing an action instead of the action itself. We derive a novel learning rule for SNN with actor-critic architecture. Since the action chosen is no longer a deterministic function of actor neurons' output, we propose a new learning rule, which we called feedback-modulated TD-STDP, to allow better credit assignment.  

We show that our learning approach can be used to solve CartPole and LunarLander at a speed comparable to standard reinforcement learning algorithms. The possibility of using SNN with biologically inspired learning rules to solve reinforcement learning tasks efficiently can provide hints on how learning may happen in biological neural systems. For example, we show that the feedback modulation in our proposed learning rule is necessary for the agent to learn, which may point to one of the possible roles of feedback connection in the biological neural system is structural credit assignment.

A learning rule similar to ours has been developed recently, called e-prop \cite{bellec2020solution}. It is based on back-propagation through time (BPTT) and it was shown that the learning rule can solve some Atari games efficiently. E-prop and our approach has several differences: (i) We include LTD side of STDP in our learning rule; (ii) In computation of pre-synaptic traces, they used $X_j = \max(0, 1 - |\frac{V_j(t)-\theta}{\theta}|)$ while we use $X_j = 1\{V_j(t) \geq \theta\}$; (iii) we propose methods for regularization that is necessary to solve more complex tasks. 

Besides more sophisticated methods to encourage exploration, possible future work can include learning hidden representations by the model, which can be applied to pixel-based games such as Atari. \cite{diehl2015unsupervised, kheradpisheh2018stdp} has shown that STDP can be used to train SNN to extract useful features for image classification. These methods can be used to train hidden layers for both actor and critic networks in our proposed model, and point to the possibility of learning a multi-layer SNN to solve a complex reinforcement learning task without back-propagation.

Comparison between SNN and ANN in learning reinforcement learning tasks also deserves further investigation. Since neurons in biological neural systems communicate by spikes, SNNs are generally considered to be closer to the biological neural system than ANN. Are there any advantages of SNN over ANN in learning, besides low communication cost? Does SNN allow faster learning or more robust policy learnt? We hope that this work can encourage more attention paid to SNN and can lead to further advance in using SNN to solve more complex reinforcement learning tasks.

\section{Acknowledgment}
We thank Andy Barto who motivated this research and provided valuable insight and comments.

\appendix
\section{Theoretical Derivation of TD-STDP and Feedback-modulated TD-STDP}

In this section, we will derive the formulas for LTP side of both TD-STDP for critic actors and feedback-modulated TD-STDP based on learning rule from actor-critic model in reinforcement learning and straight-through estimator (STE) \cite{bengio2013estimating}.

We denote $*$ as symbol for convolution: 
\begin{equation}
(f*g)(t) := \int_{-\infty}^{\infty}f(s)g(t-s)ds
\end{equation}
and define kernel $k_\tau(t) := \exp(-\frac{t}{\tau})$ for $t \geq 0$ and $0$ else. We also denote $\hat{\delta}(t)$ as the Dirac delta function.
Then firing rate of a single neuron in (\ref{eq:9}) can be written as:
\begin{equation}
\rho_j(t) = \frac{1}{\tau_n} \int_{-\infty}^t \exp \left(-\frac{t-s}{\tau_n}\right)X_j(s)ds = \frac{1}{\tau_n} (X_j * k_{\tau_n})(t) \label{eq:A1}
\end{equation}

\subsection{Proof of Theorem \ref{thm:1}} \label{sec:A1}
\begin{proof}

In actor-critic model, TD error of critic network can be minimized by the learning rule  \cite{sutton2018reinforcement}: 
\begin{equation}
w \leftarrow w + \eta \delta_t \nabla_w \hat{V}(S_t)
\end{equation}
where $\eta$ is the learning rate. In continuous time step, it can be expressed as:
\begin{equation}
\frac{dw(t)}{dt}  = \eta \delta_t \nabla_w \hat{V}(S_t) \label{eq:A10}
\end{equation}
Using (\ref{eq:5}) and (\ref{eq:A1}), $\frac{\partial  \hat{V}(S_t)}{w_{ij}} $ in our model can be computed as (assuming the weight is connected to excitatory critic neuron $j$):
\begin{align}
\frac{\partial  \hat{V}(S_t)}{\partial w_{ij}} &= \frac{\partial}{\partial w_{ij}} \alpha \left(\frac{1}{N_e} \sum_{j=1}^{N_e} \rho^c_{e,j}(t) - \frac{1}{N_i} \sum_{j=1}^{N_i} \rho^c_{i,j}(t)\right) + \beta \\
&= \frac{\partial}{\partial w_{ij}}  \frac{\alpha}{N_e}  \rho^c_{e,j}(t) \\
&= \frac{\partial}{\partial w_{ij}}  \frac{\alpha }{N_e \tau_n}  (X_j * k_{\tau_n})(t) \\
&= \frac{\alpha}{N_e \tau_n}\int_{-\infty}^t \exp\left(-\frac{t-s}{\tau_n}\right)\frac{\partial X_j(s)}{\partial w_{ij}}ds\\
&= \frac{\alpha}{N_e \tau_n} \int_{-\infty}^t \exp\left(-\frac{t-s}{\tau_n}\right) \hat{\delta}(0)  \frac{\partial I\{V_j(s) > \theta\}} {\partial w_{ij}} ds
\end{align}
We note that $\frac{\partial X_j(s)}{\partial w_{ij}}$ is zero almost everywhere. To go around the problem, we used straight-through estimator (STE) \cite{bengio2013estimating}, which was firstly proposed by Hinton in his lecture \cite{Hinton2012}. The idea is to replace the derivative of threshold function with a related surrogate. Theoretical justifications of STE and discussion on choices of surrogate can be found in \cite{yin2019understanding}. Here we used the ReLu function as our choice of surrogate. That is, we used $\frac{\partial \max(V_j(s)-\theta, 0)}{\partial w_{ij}}$ to replace $\frac{\partial I\{V_j(s) > \theta\}}{\partial w_{ij}}$. 

Also, we note that LIF model given by (\ref{eq:s0}), combined with the resetting dynamic after firing, can be written as:
\begin{align}
V_j(t) &= E_{res} + \frac{R}{\tau}\int_{-\infty}^t \exp\left(-\frac{t-s}{\tau}\right) \sum_i w_{ij}X_i(s)ds \nonumber \\ &- (\theta - E_{res}) \int_{-\infty}^t \exp\left(-\frac{t-s}{\tau}\right)  \sum_{f} \hat{\delta}(s-t^{(f)}_j) ds
\end{align}
where $\{t^{(f)}_j\} = \{t | V_j(t) > \theta\}$, the set of firing times of neuron $j$. Ignoring the effects of $w_{ij}$ on $t^{(f)}_j$, we have, 
\begin{align}
\frac{\partial V_j(s)}{\partial w_{ij}} &\approx \frac{R}{\tau} \int_{-\infty}^s \exp\left(-\frac{s-q}{\tau}\right) X_i(q)dq \\
&= \frac{R}{\tau} (X_i * k_{\tau})(s) \label{eq:A2}
\end{align}
Thus combining STE and (\ref{eq:A2}), we obtain:
\begin{align}
\frac{\partial \hat{V}(S_t)}{\partial w_{ij}} &= \frac{\alpha }{N_e \tau_n} \int_{-\infty}^t \exp\left(-\frac{t-s}{\tau_n}\right) \hat{\delta}(0) \frac{\partial I\{V_j(s) > \theta\}}{\partial w_{ij}}ds\\
&\approx \frac{\alpha}{N_e \tau_n} \int_{-\infty}^t \exp\left(-\frac{t-s}{\tau_n}\right)\hat{\delta}(0) \frac{\partial \max(V_j(s)-\theta, 0)}{\partial w_{ij}}ds \\
&= \frac{ \alpha }{N_e \tau_n} \int_{-\infty}^t \exp\left(-\frac{t-s}{\tau_n}\right)X_j(s)\frac{\partial V_j(s)}{\partial w_{ij}}ds \\
&\approx \frac{R\alpha }{N_e \tau_n \tau} \int_{-\infty}^t \exp\left(-\frac{t-s}{\tau_n}\right)X_j(s)(X_i * k_{\tau})(s) ds \\
&= \frac{R \alpha}{N_e \tau_n \tau} ((X_j \cdot (X_i * k_{\tau})) * k_{\tau_n})(t) 
\end{align}
Substituting back into (\ref{eq:A10}) and absorbing  $\frac{R\alpha }{N_e \tau_n \tau}$ into learning rate $\eta$, the learning rule is therefore given by:
\begin{equation}
\frac{dw_{ij}(t)}{dt} = \eta \delta(t) ((X_j \cdot (X_i * k_{\tau})) * k_{\tau_n})(t)  \label{eq:A3}
\end{equation}
Using the fact that the differential equation $\frac{dx(t)}{dt} = -\frac{x(t)}{\tau}+m(t)$ has solution $x(t) = (m * k_\tau)(t)$, the learning rule for critic neurons given by (\ref{eq:8}) to (\ref{eq:8E}) is equivalent to that in (\ref{eq:A3}) with $A_+ = 1$, $A_- = 0$, $\tau_p = \tau$ and $\tau_z = \tau_n$. The derivation for learning rule of inhibitory critic neurons is similar but with an opposite sign.

\end{proof}

\subsection{Proof of Theorem \ref{thm:2}} \label{sec:A2}
\begin{proof}

In actor critic model with eligibility trace, actor network can be trained by the following learning rule \cite{sutton2018reinforcement}: 
\begin{align}
q &\leftarrow \gamma \lambda q +  \nabla_w \log P(A_t|S_t) \\
w &\leftarrow w + \eta \delta_t q
\end{align}
where $\eta$ is the learning rate, $\gamma$ is the discount rate and $\lambda$ is the trace discount rate. In continuous time step, it can be expressed as:
\begin{align}
\frac{dq(t)}{dt} &= -(1- \gamma \lambda ) q(t) +  \nabla_w \log P(A(t)|S(t))  \label{eq:A11} \\
\frac{dw(t)}{dt} &= \eta \delta(t) q(t) \label{eq:A12}
\end{align}
Denote $h(t) =  \nabla_w \log P(A(t)|S(t))$, then $q(t)$ can be expressed as $(h * k_{\tau_h})(t)$ where $\tau_h = \frac{1}{1 - \gamma \lambda}$. 

We can compute $\frac{\partial \log P(A(t)|S(t)) }{\partial w_{ij}}$ as follows, using formula (\ref{eq:AN_1}) to (\ref{eq:7}) (assuming the weight is connected to excitatory actor neuron $j$ for action $k$):
\begin{align}
\frac{\partial \log P(A(t)|S(t)) }{\partial w_{ij}}&= \frac{\partial}{\partial w_{ij}} \log [\text{softmax} (\alpha(\rho^a_e (t)- \rho^a_i (t)))]_{A(t)} \\
&= \alpha(A_k(t) - s_k(t))  \frac{\partial}{\partial w_{ij}} \rho^a_{e, k} (t)\\
&= \alpha(A_k(t) - s_k(t))  \frac{\partial}{\partial w_{ij}}  \frac{1}{N_e} \sum_{j=1}^{N_e} \rho^a_{e, k, j}(t)\\
&= \frac{\alpha}{N_e}  (A_k(t) - s_k(t))  \frac{\partial}{\partial w_{ij}}  \rho^a_{e, k, j}(t)\\
&= \frac{\alpha }{N_e \tau_n} (A_k(t) - s_k(t)) \frac{\partial}{w_{ij}} (X_j * k_{\tau_n})(t) \\
&= \frac{\alpha}{N_e \tau_n} (A_k(t) - s_k(t)) \int_{-\infty}^t \exp\left(-\frac{t-s}{\tau_n}\right)\frac{\partial X_j(s)}{\partial w_{ij}}ds
\end{align}
Again, $\frac{\partial X_j(s)}{\partial w_{ij}}$ is almost zero everywhere and we use the same technique in derivation of TD-STDP to go around the problem. That is, we use $\frac{\partial X_j(s)}{\partial w_{ij}} \approx \frac{R}{\tau} X_j(s)(X_i * k_{\tau})(s)$. Then, it follows:
\begin{align}
\frac{\partial \log P(A(t)|S(t)) }{\partial w_{ij}}&= \frac{\alpha }{N_e \tau_n} (A_k(t) - s_k(t)) \int_{-\infty}^t \exp\left(-\frac{t-s}{\tau_n}\right)\frac{\partial X_j(s)}{\partial w_{ij}}ds \\
&\approx \frac{R \alpha }{N_e \tau_n \tau} (A_k(t) - s_k(t)) \int_{-\infty}^t \exp\left(-\frac{t-s}{\tau_n}\right) X_j(s)(X_i * k_{\tau})(s)ds \\
&=  \frac{R \alpha }{N_e \tau_n \tau} (A_k(t) - s_k(t))((X_j \cdot (X_i * k_{\tau})) * k_{\tau_n})(t) \\
&=  \frac{R \alpha }{N_e  \tau_n\tau} ((A_k - s_k) \cdot ((X_j \cdot (X_i * k_{\tau})) * k_{\tau_n})) (t)
\end{align}
Substituting back into $q(t)$, we thus obtain:
\begin{align}
q(t) &= (h * k_{\tau_h})(t) \\
&=  \frac{R \alpha}{N_e  \tau_n \tau}  (((A_k - s_k) \cdot ((X_j \cdot (X_i * k_{\tau})) * k_{\tau_n})) * k_{\tau_h})(t)
\end{align}
Substituting back into (\ref{eq:A12}) and absorbing  $\frac{R \alpha  }{N_e \tau_n \tau}$ into learning rate $\eta$, the learning rule is therefore given by:
\begin{equation}
\frac{dw_{ij}(t)}{dt} = \eta \delta(t) (((A_k - s_k) \cdot ((X_j \cdot (X_i * k_{\tau})) * k_{\tau_n})) * k_{\tau_h})(t) \label{eq:A4}
\end{equation}

Using the fact that the differential equation $\frac{dx(t)}{dt} = -\frac{x(t)}{\tau}+m(t)$ has solution $x(t) = (m * k_\tau)(t)$, the learning rule for actor neurons given by (\ref{eq:6}) to (\ref{eq:6E}) is equivalent to that in (\ref{eq:A4}) with $A_+ = 1$, $A_- = 0$, $\tau_p = \tau$, $\tau_z = \tau_n$, and $\tau_q = \tau_h$. The derivation for learning rule of inhibitory actor neuron is similar but with an opposite sign.
\end{proof}

\section{Formulas in Discrete Time Step}

The computation of TD error in (2) and learning rules in Section 3 are all in differential form. Since we simulate SNN using discrete time steps, we will list the corresponding formulas in discrete time step here for completeness.

Firstly, we can estimate the TD error from $t$ to $t + \Delta t$ by integrating both side of (\ref{eq:0}). Using the fact that $\exp(-\frac{-\Delta t}{\tau}) \approx 1-\frac{\Delta t}{\tau}$, we have:

\begin{equation}
\delta_{t:t+\Delta t} \approx \exp \left(\frac{-\Delta t}{\tau} \right)\hat{V}(S_{t+\Delta t})+ R_t \Delta t - \hat{V}(S_{t})\label{eq:D1}
\end{equation}

However, this equation assumes $R_t$ does not decay for the period $\Delta t$ when used to estimate the value of $V(S_t)$. A better estimate is to adjust for this decay as well, by using $\frac{\Delta t}{2}$ as duration for discount:

\begin{equation}
\delta_{t:t+\Delta t} \approx  \exp\left(\frac{-\Delta t}{\tau}\right)\hat{V}(S_{t+\Delta t})+ \exp\left(\frac{-\Delta t}{2\tau}\right) R_t \Delta t - \hat{V}(S_{t}) \label{eq:A13}
\end{equation}

The equation in (\ref{eq:A13}) is intuitive. The discounted return from $t$ to $t+\Delta t$ is approximated by $\exp\left(\frac{-\Delta t}{2\tau}\right) R_t$ while the discounted return from $t+\Delta t$ onward is approximated by $\exp(\frac{-\Delta t}{\tau})\hat{V}(S_{t+\Delta t})$. (\ref{eq:A13}) will be used to compute TD error $\delta$ in (\ref{eq:A6}) and (\ref{eq:A7}). If the episode ended already, then $\hat{V}(S_{t+\Delta t})$ in (\ref{eq:A13}) is replaced with $0$.

For learning rule, we denote $\hat{X}_i$ as the indicator function of whether neuron $i$ is firing, equivalent to re-scaling the Dirac delta functions in spike train $X_i$ to 1. Then, the learning rule of weight from input neuron $i$ to critic neuron $j$ in discrete time step is as follows (corresponds to (\ref{eq:8}) to (\ref{eq:8E})):
\begin{align}
P_{j} &\leftarrow \exp\left(\frac{-\Delta t}{\tau_p}\right)P_{j} + \hat{X}_j \\
P_{i} &\leftarrow \exp\left(\frac{-\Delta t}{\tau_p}\right)P_{i} + \hat{X}_i \\
z_{ij} &\leftarrow \exp\left(\frac{-\Delta t}{\tau_z}\right)z_{ij} + A_+ P_{i} \hat{X}_j - A_- P_{j} \hat{X}_i  \\
w_{ij} &\leftarrow w_{ij} \pm \eta \delta z_{ij} \label{eq:A6}
\end{align}

The learning rule for weight from input neuron $i$ to actor neuron $j$ (denote $k$ as the action for this actor neuron) in discrete time step is as follows (corresponds to (\ref{eq:6}) to (\ref{eq:6E})):

\begin{align}
P_{j} &\leftarrow \exp\left(\frac{-\Delta t}{\tau_p}\right)P_{j} + \hat{X}_j \\
P_{i} &\leftarrow \exp\left(\frac{-\Delta t}{\tau_p}\right)P_{i} + \hat{X}_i \\
z_{ij} &\leftarrow \exp\left(\frac{-\Delta t}{\tau_z}\right)z_{ij} + A_+ P_{i} \hat{X}_j - A_- P_{j} \hat{X}_i  \\
q_{ij} &\leftarrow \exp\left(\frac{-\Delta t}{\tau_q}\right)q_{ij} + (A_k - s_k)z_{ij} \\
w_{ij} &\leftarrow w_{ij} \pm \eta \delta q_{ij} \label{eq:A7}
\end{align}

In (\ref{eq:A6}) and (\ref{eq:A7}), excitatory neurons are updated with positive sign while inhibitory neurons are updated with negative sign.

\bibliographystyle{unsrt}
\bibliography{citation}

\begin{thebibliography}{10}

\bibitem{pfeiffer2018deep}
Michael Pfeiffer and Thomas Pfeil.
\newblock Deep learning with spiking neurons: opportunities and challenges.
\newblock {\em Frontiers in neuroscience}, 12:774, 2018.

\bibitem{huh2018gradient}
Dongsung Huh and Terrence~J Sejnowski.
\newblock Gradient descent for spiking neural networks.
\newblock In {\em Advances in Neural Information Processing Systems}, pages
  1433--1443, 2018.

\bibitem{zhang2019spike}
Wenrui Zhang and Peng Li.
\newblock Spike-train level backpropagation for training deep recurrent spiking
  neural networks.
\newblock In {\em Advances in Neural Information Processing Systems}, pages
  7800--7811, 2019.

\bibitem{obeid2019structured}
Dina Obeid, Hugo Ramambason, and Cengiz Pehlevan.
\newblock Structured and deep similarity matching via structured and deep
  hebbian networks.
\newblock In {\em Advances in Neural Information Processing Systems}, pages
  15377--15386, 2019.

\bibitem{dayan2001theoretical}
Peter Dayan and Laurence~F Abbott.
\newblock Theoretical neuroscience: computational and mathematical modeling of
  neural systems.
\newblock 2001.

\bibitem{hassabis2017neuroscience}
Demis Hassabis, Dharshan Kumaran, Christopher Summerfield, and Matthew
  Botvinick.
\newblock Neuroscience-inspired artificial intelligence.
\newblock {\em Neuron}, 95(2):245--258, 2017.

\bibitem{gerstner2014neuronal}
Wulfram Gerstner, Werner~M Kistler, Richard Naud, and Liam Paninski.
\newblock {\em Neuronal dynamics: From single neurons to networks and models of
  cognition}.
\newblock Cambridge University Press, 2014.

\bibitem{toyoizumi2005spike}
Taro Toyoizumi, Jean-Pascal Pfister, Kazuyuki Aihara, and Wulfram Gerstner.
\newblock Spike-timing dependent plasticity and mutual information maximization
  for a spiking neuron model.
\newblock In {\em Advances in neural information processing systems}, pages
  1409--1416, 2005.

\bibitem{florian2007reinforcement}
R{\u{a}}zvan~V Florian.
\newblock Reinforcement learning through modulation of spike-timing-dependent
  synaptic plasticity.
\newblock {\em Neural Computation}, 19(6):1468--1502, 2007.

\bibitem{fremaux2013reinforcement}
Nicolas Fr{\'e}maux, Henning Sprekeler, and Wulfram Gerstner.
\newblock Reinforcement learning using a continuous time actor-critic framework
  with spiking neurons.
\newblock {\em PLoS computational biology}, 9(4), 2013.

\bibitem{vasilaki2009spike}
Eleni Vasilaki, Nicolas Fr{\'e}maux, Robert Urbanczik, Walter Senn, and Wulfram
  Gerstner.
\newblock Spike-based reinforcement learning in continuous state and action
  space: when policy gradient methods fail.
\newblock {\em PLoS Comput Biol}, 5(12):e1000586, 2009.

\bibitem{bing2018end}
Zhenshan Bing, Claus Meschede, Kai Huang, Guang Chen, Florian Rohrbein, Mahmoud
  Akl, and Alois Knoll.
\newblock End to end learning of spiking neural network based on r-stdp for a
  lane keeping vehicle.
\newblock In {\em 2018 IEEE International Conference on Robotics and Automation
  (ICRA)}, pages 1--8. IEEE, 2018.

\bibitem{aenugu2019reinforcement}
Sneha Aenugu, Abhishek Sharma, Sasikiran Yelamarthi, Hananel Hazan, Philip~S.
  Thomas, and Robert Kozma.
\newblock Reinforcement learning with a network of spiking agents, 2019.

\bibitem{potjans2009spiking}
Wiebke Potjans, Abigail Morrison, and Markus Diesmann.
\newblock A spiking neural network model of an actor-critic learning agent.
\newblock {\em Neural computation}, 21(2):301--339, 2009.

\bibitem{wunderlich2019demonstrating}
Timo Wunderlich, Akos~F Kungl, Eric M{\"u}ller, Andreas Hartel, Yannik
  Stradmann, Syed~Ahmed Aamir, Andreas Gr{\"u}bl, Arthur Heimbrecht, Korbinian
  Schreiber, David St{\"o}ckel, et~al.
\newblock Demonstrating advantages of neuromorphic computation: a pilot study.
\newblock {\em Frontiers in neuroscience}, 13:260, 2019.

\bibitem{witten1977adaptive}
Ian~H Witten.
\newblock An adaptive optimal controller for discrete-time markov environments.
\newblock {\em Information and control}, 34(4):286--295, 1977.

\bibitem{barto1983neuronlike}
Andrew~G Barto, Richard~S Sutton, and Charles~W Anderson.
\newblock Neuronlike adaptive elements that can solve difficult learning
  control problems.
\newblock {\em IEEE transactions on systems, man, and cybernetics},
  (5):834--846, 1983.

\bibitem{takahashi2008silencing}
Yuji Takahashi, Geoffrey Schoenbaum, and Yael Niv.
\newblock Silencing the critics: understanding the effects of cocaine
  sensitization on dorsolateral and ventral striatum in the context of an
  actor/critic model.
\newblock {\em Frontiers in neuroscience}, 2:14, 2008.

\bibitem{glimcher2011understanding}
Paul~W Glimcher.
\newblock Understanding dopamine and reinforcement learning: the dopamine
  reward prediction error hypothesis.
\newblock {\em Proceedings of the National Academy of Sciences}, 108(Supplement
  3):15647--15654, 2011.

\bibitem{roelfsema2018control}
Pieter~R Roelfsema and Anthony Holtmaat.
\newblock Control of synaptic plasticity in deep cortical networks.
\newblock {\em Nature Reviews Neuroscience}, 19(3):166, 2018.

\bibitem{roelfsema2005attention}
Pieter~R Roelfsema and Arjen~van Ooyen.
\newblock Attention-gated reinforcement learning of internal representations
  for classification.
\newblock {\em Neural computation}, 17(10):2176--2214, 2005.

\bibitem{rombouts2015attention}
Jaldert~O Rombouts, Sander~M Bohte, and Pieter~R Roelfsema.
\newblock How attention can create synaptic tags for the learning of working
  memories in sequential tasks.
\newblock {\em PLoS Comput Biol}, 11(3):e1004060, 2015.

\bibitem{bellec2020solution}
Guillaume Bellec, Franz Scherr, Anand Subramoney, Elias Hajek, Darjan Salaj,
  Robert Legenstein, and Wolfgang Maass.
\newblock A solution to the learning dilemma for recurrent networks of spiking
  neurons.
\newblock {\em Nature communications}, 11:3625, 2020.

\bibitem{guo2009continuous}
Xianping Guo and On{\'e}simo Hern{\'a}ndez-Lerma.
\newblock Continuous-time markov decision processes.
\newblock In {\em Continuous-Time Markov Decision Processes}, pages 9--18.
  Springer, 2009.

\bibitem{sutton2018reinforcement}
Richard~S Sutton and Andrew~G Barto.
\newblock {\em Reinforcement learning: An introduction}.
\newblock MIT press, 2018.

\bibitem{bengio2013estimating}
Yoshua Bengio, Nicholas L{\'e}onard, and Aaron Courville.
\newblock Estimating or propagating gradients through stochastic neurons for
  conditional computation.
\newblock {\em arXiv preprint arXiv:1308.3432}, 2013.

\bibitem{kingma2014adam}
Diederik~P Kingma and Jimmy Ba.
\newblock Adam: A method for stochastic optimization.
\newblock {\em arXiv preprint arXiv:1412.6980}, 2014.

\bibitem{mnih2016asynchronous}
Volodymyr Mnih, Adria~Puigdomenech Badia, Mehdi Mirza, Alex Graves, Timothy
  Lillicrap, Tim Harley, David Silver, and Koray Kavukcuoglu.
\newblock Asynchronous methods for deep reinforcement learning.
\newblock In {\em International conference on machine learning}, pages
  1928--1937, 2016.

\bibitem{Hazan2018}
Hananel Hazan, Daniel~J. Saunders, Hassaan Khan, Devdhar Patel, Darpan~T.
  Sanghavi, Hava~T. Siegelmann, and Robert Kozma.
\newblock Bindsnet: A machine learning-oriented spiking neural networks library
  in python.
\newblock {\em Frontiers in Neuroinformatics}, 12:89, 2018.

\bibitem{konidaris2011value}
George Konidaris, Sarah Osentoski, and Philip Thomas.
\newblock Value function approximation in reinforcement learning using the
  fourier basis.
\newblock In {\em Twenty-fifth AAAI conference on artificial intelligence},
  2011.

\bibitem{brockman2016openai}
Greg Brockman, Vicki Cheung, Ludwig Pettersson, Jonas Schneider, John Schulman,
  Jie Tang, and Wojciech Zaremba.
\newblock Openai gym.
\newblock {\em arXiv preprint arXiv:1606.01540}, 2016.

\bibitem{diehl2015unsupervised}
Peter~U Diehl and Matthew Cook.
\newblock Unsupervised learning of digit recognition using
  spike-timing-dependent plasticity.
\newblock {\em Frontiers in computational neuroscience}, 9:99, 2015.

\bibitem{kheradpisheh2018stdp}
Saeed~Reza Kheradpisheh, Mohammad Ganjtabesh, Simon~J Thorpe, and Timoth{\'e}e
  Masquelier.
\newblock Stdp-based spiking deep convolutional neural networks for object
  recognition.
\newblock {\em Neural Networks}, 99:56--67, 2018.

\bibitem{Hinton2012}
G.~Hinton.
\newblock Neural networks for machine learning. coursera, video lectures.
  lecture 9c.
\newblock
  \url{https://www.youtube.com/watch?v=LN0xtUuJsEI&list=PLoRl3Ht4JOcdU872GhiYWf6jwrk_SNhz9&index=41},
  2012.

\bibitem{yin2019understanding}
Penghang Yin, Jiancheng Lyu, Shuai Zhang, Stanley Osher, Yingyong Qi, and Jack
  Xin.
\newblock Understanding straight-through estimator in training activation
  quantized neural nets.
\newblock {\em arXiv preprint arXiv:1903.05662}, 2019.

\end{thebibliography}

\end{document}